\documentclass{article}

\usepackage[affil-it]{authblk}
\usepackage{url}
\usepackage{footnote}
\usepackage{amsmath}
\usepackage{amsfonts}
\usepackage{booktabs} 
\usepackage{color}
\usepackage{multicol}
\usepackage{enumitem}
\usepackage{bm}
\usepackage[normalem]{ulem}
\usepackage{multirow}
\usepackage{pifont}
\usepackage[linesnumbered,ruled, noend]{algorithm2e}
\usepackage{graphicx}
\usepackage[outdir=./]{epstopdf}

\usepackage{natbib} 
\def\tm{T_{\mathcal{G}}}
 
\usepackage{subcaption}

\usepackage{scalerel,stackengine}
\stackMath
\newcommand\reallywidehat[1]{%
\savestack{\tmpbox}{\stretchto{%
  \scaleto{%
    \scalerel*[\widthof{\ensuremath{#1}}]{\kern.1pt\mathchar"0362\kern.1pt}%
    {\rule{0ex}{\textheight}}
  }{\textheight}%
}{2.4ex}}%
\stackon[-6.9pt]{#1}{\tmpbox}%
}

\AtBeginDocument{%
  \providecommand\BibTeX{{%
    \normalfont B\kern-0.5em{\scshape i\kern-0.25em b}\kern-0.8em\TeX}}}

\begin{document}

\title{HTTE: A Hybrid Technique For Travel Time Estimation In Sparse Data Environments}

\date{}
\maketitle

\author{Nikolaos Zygouras$^1$, \thanks{zygouras@di.uoa.gr}}
\author{Nikolaos Panagiotou$^1$, \thanks{npanagio@di.uoa.gr}}
\author{Yang Li$^3$, \thanks{yangli@sz.tsinghua.edu.cn}}
\author{Dimitrios Gunopulos$^1$, \thanks{dg@di.uoa.gr}}
\author{Leonidas Guibas$^2$, \thanks{guibas@cs.stanford.edu}}

\footnote{(1)National and Kapodistrian University of Athens (2)Stanford University
(3) Tsinghua-Berkeley Shenzhen Institute}


\begin{abstract}
\textit{Travel time estimation}  is a critical task, useful to many urban
applications at the individual citizen and the stakeholder level.
This paper presents a novel hybrid algorithm for travel time estimation
that leverages historical and sparse real-time trajectory data.
Given a path and a departure time we estimate the travel time
taking into account the historical information, the real-time
trajectory data and the correlations among different road segments.
We detect similar road segments using historical trajectories,
and use a latent representation to model the similarities. Our
experimental evaluation demonstrates the effectiveness of our approach.
\end{abstract}







%


%


\section{Introduction}

The increasing population density in modern cities is leading to massively increasing commuting demands for citizens.
This strongly motivates the need for faster and more efficient navigation tools in the city. To be truly useful such systems need to be able to monitor and  accurately predict the traffic conditions across the entire city road network in real-time, to respond to abrupt or  unexpected condition changes.
Accurate  {\it travel time estimation} for a path in the road network is important for tools that help individual citizens plan their travel; equally stakeholders and city/traffic authorities can exploit such tools for efficient route planning and automatic detection of traffic anomalies.
Several works have used data from static sensors, including loop detectors~\cite{kwon2003estimation} and CCTV cameras~\cite{zhan2015lane} to address the travel time estimation for a path. Such sensors are typically located at several junctions across the city monitoring the traffic condition. 
The prevalence of such solutions has diminished since their first appearance. The increased capital cost of installing and maintaining such devices and their limited and static coverage of the road network in combination with the inherent inaccuracy in calculating the travel time from the output of these sensors (i.e. number of vehicles, speed and video frames) limit their application in practice.

Recently researchers have used trajectory data in order to perform travel time estimation, thus taking advantage of the widespread use of mobile devices that are equipped with Global Position System (GPS) technology. 
Thus, such mobile devices are transformed into important moving and  ubiquitous sensors reporting the traffic condition at different parts of the road network.

However, not all such data are available in real-time for a variety of reasons. Sensors may be offline or smartphones may report their locations  infrequently or  in batches. Additionally, in several cases the application has  access to a small number of  sensors. Taxi or bus companies, for instance, have data  for the vehicles in their fleet only. Over time, such applications can compile massive historical data with impressive coverage, although at any given time the coverage of the map is sparse.


The \emph{goal of this study} is to estimate the time that is required to travel a given query path considering a particular departure time in real time even when very patchy real time coverage of the network is available.
To accomplish this we propose a hybrid model that  considers efficiently  the recent and historical trajectories generated by a sample of vehicles.
In our settings, the travel time estimation is a challenging problem for the following reasons:

\noindent \emph{1. Data sparsity:} for the majority of the road segments we do not have any information regarding their recent traffic condition, since only the trajectories of a small subset of vehicles moving in the road network is available. Therefore, our setting is different from industry situations where an extensive real-time coverage of  the  traffic conditions may be available.

\noindent \emph{2. Noisy data:} the travel time reports  are extremely noisy. A driver may stop at a traffic light and spend a while waiting, while another driver crosses the junction without stopping at the traffic light. This  would generate two divergent travel time reports for the  same road segment. Also, some drivers may drive faster or slower than others adding further complexity in the measurements.

\noindent \emph{3. Unpredictable dynamics of traffic:} there are many traffic irregularities and anomalies that may occur in the road network (i.e. an accident, a social event etc.) that could affect the expected and the orderly traffic behaviour of the road network.

\noindent \emph{4. Response time:} it is crucial to create a model able to answer users' queries instantly and at the same time update its state in real time considering the recent traffic condition.


To address the aforementioned challenges, we propose a \underline{H}ybrid \underline{T}ravel \underline{T}ime \underline{E}stimation framework, referred as HTTE. The framework achieves the estimation of the travel time for a given query path (laying on top of the road network), using data from moving vehicles.
The proposed framework is capable of providing predictions in real-time by exploiting the similarity of the road segments and by considering travel time  reports provided by recent as  well as  historical trajectories. 
The contributions of this work can be summarized as follows:

\begin{itemize}[leftmargin=*]
    \item \emph{A latent representation for road segments:} In order to treat the  data sparsity problem in individual segments, we take advantage of the available traffic information from other segments with similar traffic behavior. We provide a mechanism for learning a latent representation for the road segments. This representation describes their traffic behaviour. Thus, road segments with similar traffic behaviour will be placed close in this latent space. 
    
    
    \item \emph{A Hybrid Estimation Model:} We develop a streaming and hybrid estimation model that  captures the recent traffic reports, the periodicity of the time series and the correlations among different road segments. Our framework models large areas of the city jointly and not the road segments individually, addressing this way the data sparsity problem.
    \item We \emph{evaluate} our method under realistic settings using data from buses and taxis and compare it with state of the art techniques. 
    We show that hybrid techniques such as the one we propose outperform techniques that use only historical or only real-time and near real-time data.
    Experiments show that incorporating pathlets can  improve the query efficiency up to 14 times with slight degradation in accuracy. This modification allows HTTE to work with  interactive applications on a much larger scale.

\end{itemize}


\section{Related Work}
\label{sec:related}

\noindent
{\bf Travel Time Estimation Using Static Sensor Data:}
A variety of techniques have been proposed in the literature for estimating the road segments' traffic flow or speed, exploiting static sensor data. Among these techniques, \cite{qu2008bpca} describes a matrix decomposition method, for estimating the traffic flow  in  Beijing, \cite{li2013efficient} proposes an extension of Probabilistic Principal Component Analysis and Kernel Principal Component Analysis that captures spatial and temporal dependencies and \cite{wang2016traffic} uses deep convolutional neural networks for predicting the road segments' speed.
A DeepNN architecture able to capture spatial/temporal relations between road segments was proposed in~\cite{DBLP:conf/iclr/LiYS018} using speeds from static sensors.





\noindent
{\bf Travel Time Estimation Using Dynamic Mobile Sensor Data:}
Many studies have explored the travel time estimation problem using moving sensors.
In \cite{zhan2013urban} the  authors estimated the travel time in links, employing least-square optimization on taxi trip data that contained endpoint locations and trip metadata. 
A Bayesian mixture model was introduced in~\cite{zhan2016bayesian} that estimated the short-term average urban link travel times with partial information available.
The correlations between the travel times of nearby links and different time slots are crucial for inferring the traffic state
of a particular link \cite{niu2014deepsense,zhang2016urban}.
Online methods that determine the time required by a bus to reach a specified bus stop were proposed in~\cite{gal2015traveling}, \cite{Gal2018} and \cite{yu2011bus}. 
In \cite{wang2016simple} the authors 
propose a method that estimates the travel time by identifying near-neighbor trajectories,
with similar origin and destination.
The final estimation of the travel time is the weighted average of the neighbors travel times.
The authors in~\cite{jenelius2013travel} state that the travel time  estimation  can be approximated by the sum of the segments' traversal time and a delay penalty that occurs at the links between the segments. 
In~\cite{citywidetraffic} the authors 
propose a hybrid framework that incorporates (i) road network data, (ii) POI (Points of Interest), (iii) GPS trajectories and (iv) weather information to estimate the \textit{travel speed} and the \textit{traffic volume}.
In~\cite{li2017urban} the authors proposed a technique that estimates the travel time 
using a small number of GPS-equipped   cars available, discovering local traffic patterns over a set of frequent paths, a.k. a. pathlets.
At query time, a trajectory is decomposed into pathlets, whose recent travel time is estimated using pattern matching with recent travel time observations.
A spatio-temporal hidden Markov model that models correlations among different traffic time series was proposed in
\cite{DBLP:journals/pvldb/0002GJ13}
taking into account the sparsity, the spatio-temporal correlation, and the
heterogeneity of time series.
A different approach was followed by \cite{DBLP:journals/vldb/YangDGJH18,DBLP:journals/pvldb/DaiYGJH16} assigning the weights to the paths instead of the edges of the road network, avoiding splitting the trajectories in small fragments.
The authors in~\cite{DBLP:journals/tkde/YangKJ14} explored the use of weighted PageRank values of edges for assigning appropriate weights to all edges.
The authors in \cite{DBLP:conf/aaai/IdeS11} proposed a weight propagation model able to capture neighboring road-link dependencies and embedded the model to the regression task. In the same direction in \cite{DBLP:conf/aaai/ZhengN13} the authors provided a multi task learning framework that simultaneously captures spatial dependencies and temporal dynamics encouraging spatio-temporal smoothness.




\noindent
{\bf Learning Latent Features on the Road Network:}
Recent techniques have suggested more sophisticated methods for taking advantage of historical data for travel time prediction. 
In \cite{hofleitner2012learning}  a technique that estimates the arterial travel time distributions is proposed, introducing hidden random variables that represent the road segments' state (congested and undersaturated). Then a dynamic bayesian network learns the travel time distributions.
A technique that detects the time-varying distribution of travel time of road segments using Graph Convolutional Neural Network was introduced in~\cite{DBLP:conf/icde/HuG0J19}.
In~\cite{deng2016latent} the authors proposed a method that imputes the short future speeds for the road segments,
utilizing latent topological and temporal features learned and updated incrementally through matrix factorization.
In~\cite{wang2014travel} the travel times of different road segments, drivers and time slots are modeled as a 3D sparse tensor. The missing values were filled in using the geospatial features 
and the recent and historical traffic information. A dynamic programming technique optimally concatenated the path into subpaths.

\noindent
{\bf Deep Learning Approaches:}
The recent success of deep learning in a variety of learning problems, lead to the design of deep learning architectures for the travel time  estimation task. 
In  DeepTravel~\cite{ijcai2018-508} a deep learning architecture is proposed with two major components. The first handles the representation of the features (spatial, temporal, driving state) with an embeddings layer while the second consists of  a BiLSTM layer that performs the actual regression. 
An origin-destination travel time estimation method is MURAT~\cite{li2018multi} that
employs a graph embedding method for extracting roads' embeddings and an embedding layer for capturing the spatial and temporal features. These embeddings layers transform and provide the input to a Residual network.
The authors in~\cite{wang2018will} proposed an end-to-end Deep learning framework for travel time estimation of an entire path (DeepTTE). A geo-convolution operation is proposed that handles the GPS points of the trajectory followed by a recurrent component. A multi-task learning component is used in order to learn both the total travel time of the given path and the travel times of smaller parts of the path.
Finally in~\cite{wang2018learning} the authors proposed a deep learning model that estimates the time of arrival using wide, deep and recurrent components.

In essence, in our work we exploit novel techniques for the discovery of latent features in the spatial and temporal traffic data and at the same time leverage the use of sparse real-time information.

\section{Our Approach}
\label{sec:our-approach}

\subsection{Problem Setup}
\label{sec:definitions}

In this work we propose an efficient algorithm for estimating the travel time that is required for a vehicle to traverse a path of the road network. The proposed framework receives firstly as input a set of vehicles' \textit{trajectories}. Each \textit{trajectory} is a sequence of time ordered spatial points $T: (p_1, t_1) \rightarrow \dots \rightarrow (p_n, t_n)$, where each point $p_i\in \mathbb{R}^2$ is the sampled GPS position and $t_i$ is the corresponding timestamp of the measurement.
Then the points of the trajectories are mapped on a \textit{Road Network}. 
A \textit{Road Network} is defined as a topological structure of a network captured by a graph $\mathcal{G}$, where the nodes of $\mathcal{G}$ correspond to a collection of road segments $r_i$ that link different urban areas together and the set of edges represent the connections between these road segments.
A \textit{map-matched trajectory}
$T_{\mathcal{G}}$
is a projection of a trajectory $T$ in the road network $\mathcal{G}$. $\,T_{\mathcal{G}}:(r_1,t_{in,1}, t_1) \rightarrow \dots \rightarrow (r_{n'},t_{in,n'}, t_{n'})$ is defined as a sequence of the visited road segments $r_i$ along with the timestamps that the vehicle entered $t_{in,i}$
and left $t_{i}$ each road segment. 
In this work we are computing the estimated travel time of a given query path, without maintaining profiles for each driver.

Each vehcile that traverses a segment $r_i$ generates a \textit{Travel Time Report} $R_i=(r_i,t_i,TT_i)$. The reports $R_i$ are available when the vehicle exits the road segment. $TT_i=t_{i}-t_{in,i}$ is the travel time required for the traversal and $t_i$ is the time when the vehicle left the road segment. The travel time reports are stored in a collection $R_H$ that is incrementally updated as new reports are provided.
Also, a \textit{Road Segment Embedding} is a mapping $E: r_i \rightarrow \mathbb{R}^D$
, that maps a road segment $r_i$ of the road network $\mathcal{G}$ to a D-dimensional latent space.
Finally, a query path $P_q:r_{q_1} \rightarrow \dots \rightarrow r_{q_m}$ is an ordered sequence of $m$ consecutive road segments of $\mathcal{G}$.

\textbf{Problem Definition.} \textit{Given a query $q$ that consists of a query path $P_q$ and a departure time $t_{dep,q}$, predict the travel time $\reallywidehat{TT_q}$ that is required for a vehicle to traverse all the road segments of $P_q$ departing at $t_{dep,q}$ using the collection of historical travel time reports $R_H$ that have been received until the time of the query.}

\begin{figure}[t]
  \centering
      \includegraphics[width=0.45\textwidth]{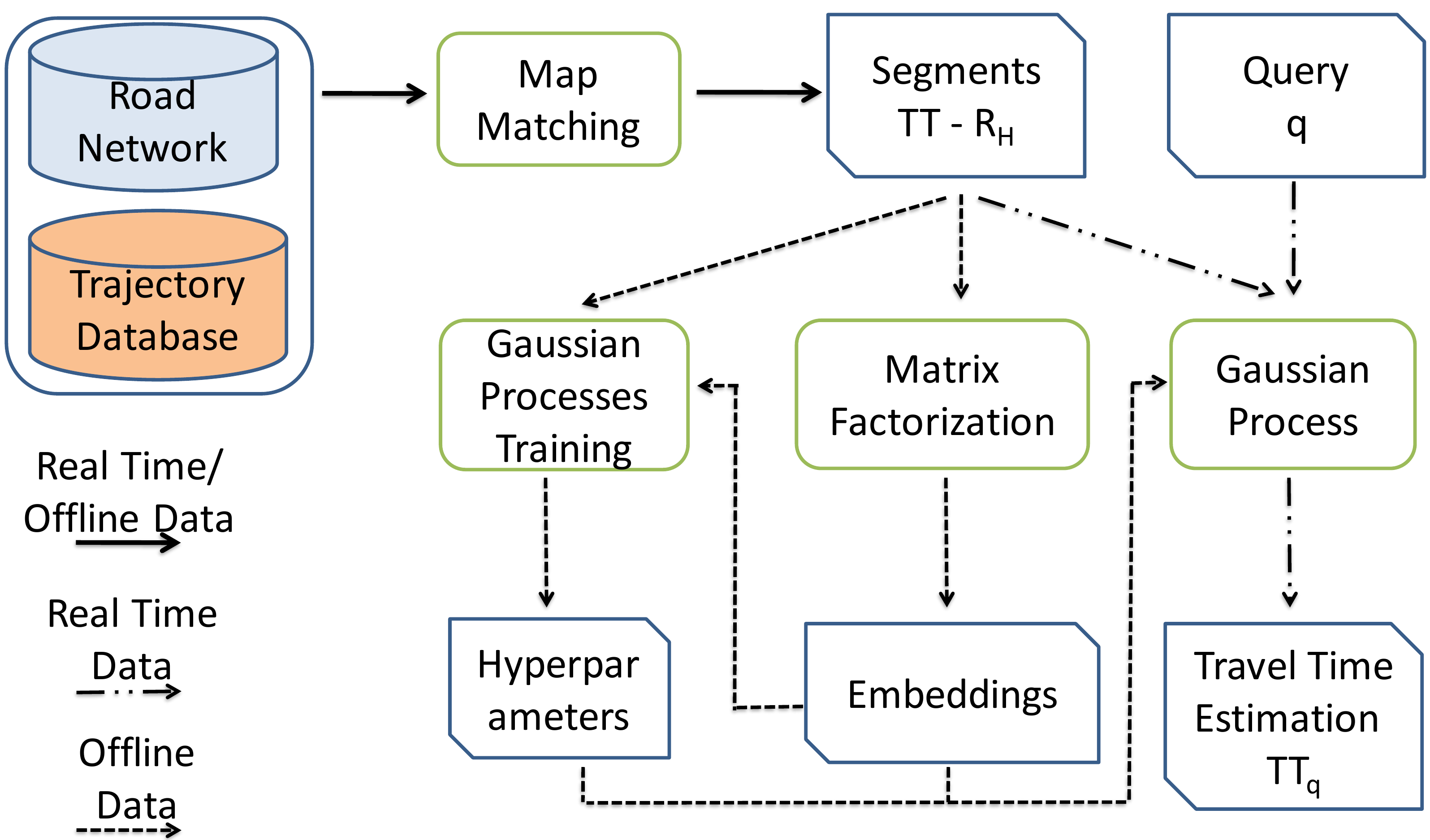}
  \caption{Framework of our approach.}
  \label{fig:our_framework}
\end{figure}

\subsection{Overview of the Approach}
\label{sec:method}
The overview of our framework for estimating the travel time of a given query path is illustrated in Figure~\ref{fig:our_framework}. Our framework has two major tasks. Initially, it aims to model the historical data by examining the traffic behavior of the road segments. Then it makes real-time predictions that exploit both historical and real-time  information.
Our architecture consists of the following modules:

\noindent\textbf{Module 1: Road Network \& Trajectory Partitioning.} The first processing component receives as input raw GPS trajectories and maps them
onto paths in a road network,
such as OpenStreetMap (OSM). 
In this case, the GPS points of a trajectory are mapped to road segments (i.e. OSM road segments) using the Barefoot\footnote{\url{https://github.com/bmwcarit/barefoot}} library.
Also, in this work we consider more abstract models of the road network. Under this case, the input GPS points could be mapped to sequences of road segments,
or to pairs of GPS locations that represent a transition from an origin to a destination. 


 
One way to obtain a compact set of road segment sequences for learning the traffic behaviour is using the concept of pathlet dictionary.
Given a set of map matched trajectories $S$, the {\bf pathlet dictionary (PD)} is a collection of paths (road segment sequences) on the road network that reconstructs all trajectories in $S$ by concatenation. Entries in the pathlet dictionary are referred as {\em pathlets}. A pathlet dictionary is considered {\em optimal} if it satisfies the following criteria:
\textit{(i)} The number of pathlets in the dictionary, $|PD|$ is minimized.
\textit{(ii)}
For each map matched trajectory $\tm \in S$, the number of pathlets used to reconstruct $\tm$, $|p(\tm)|$ is minimized.

Although computing the optimal pathlet dictionary from a trajectory collection is an NP-hard problem, \cite{chenchen2013} proposed an efficient approximation algorithm to find solution in $O(|S| \cdot n^2)$ time, where $|S|$ is the size of the trajectory collection and $n$ is the maximum number of road segments.
When the dictionary is computed, it's easy to query the decomposition of map matched trajectories using graph search creating the travel time reports for the pathlets.  



After the mapping to the road segments is completed using the former or the latter approach the travel time reports $R_H$ are generated, containing the time required to travel the road segments or the pathlets. This processing module is common for both the historical and the real-time data.
These reports are the fundamental element of the proposed travel time estimation technique.

\begin{figure}
  \centering
      \includegraphics[width=0.49\textwidth]{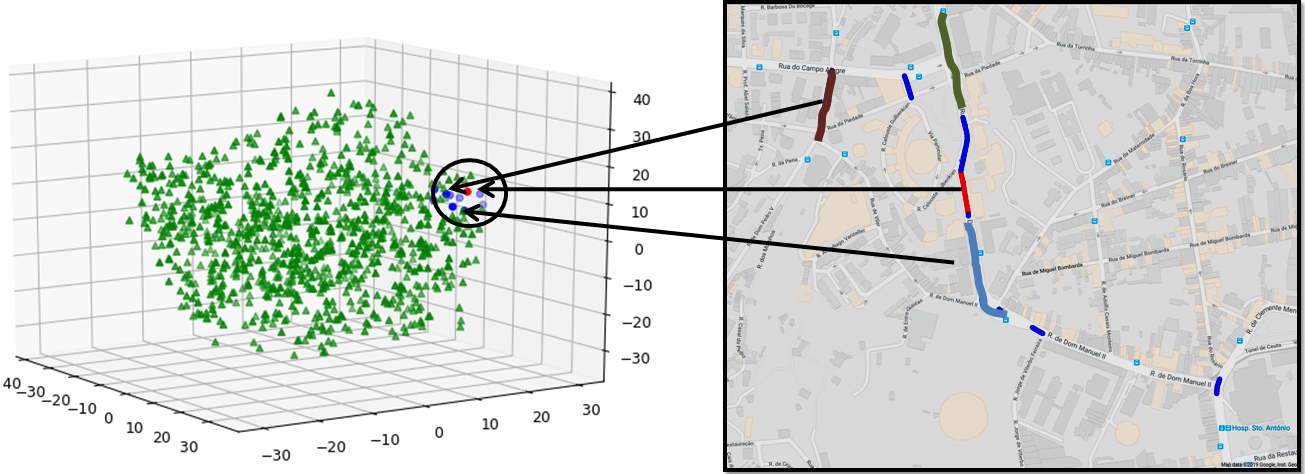}
  \caption{Example segments mapping into the embedding space. The t-SNE technique is used in order to project the embeddings in a 3-dimensional space.}
  \label{fig:embeddings}
\end{figure}

\noindent\textbf{Module 2: A latent representation for road segments. }
Identifying segments with similar traffic patterns is  crucial for tackling the data sparsity problem. This component receives as input timestamped travel time reports for the various road segments.
A latent representation for each segment is learned capturing the correlations among the road segments. That is, segments with similar traffic behaviour are placed close in the latent space. On the other hand, segments with divergent traffic behaviour are placed far apart in the latent space. Figure~\ref{fig:embeddings} on the right illustrates on the map several road segments with similar traffic behavior (similar embeddings). On the left part of the figure the mapping of these road segments into the embedding space is presented.


\noindent\textbf{Module 3: Travel Time Estimation.} The final module estimates the time needed to travel a given query path $P_q$. This module is comprised by an offline and a real-time stage.
A Gaussian Process~\citep{williams2006gaussian} model is trained offline with a set of historical reports $R_H$ using a complex covariance function, able to capture a variety of data aspects such as the data periodicity and the magnitude of the most recent values. Then our system receives in real-time queries $q$ and estimates the total travel time required to traverse the given query path $P_q$, estimating the travel times of all the individual road segments of $P_q$.




\section{THE HTTE Algorithm}
\label{sec:method}
We describe the HTTE travel time estimation algorithm.
Initially, we  describe a technique for modeling the traffic data and for identifying road segments with similar traffic behavior. Then we discuss the desirable properties of our data and present a covariance function that captures these properties. Finally, we describe a method for predicting in real-time the travel times of the query paths.

\subsection*{A Latent Representation for Road Segments}
\label{sec:embeddings}
Data sparsity is one of the major obstacles in estimating the travel time of a road segment since traffic information is provided only by a few vehicles. Information for the recent traffic state for the majority of road segments is often missing. It is essential to ensure that the method will use the traffic information from the segments for which we have recent reports in order to infer the state for a segment with similar traffic behavior but without recent information. 

Here we propose a technique that detects a latent embedding representation for the  segments. The main property of this mapping is  that segments with similar traffic behaviour should be placed close in this embedding space. This latent representation is used by the proposed prediction model in order to address the data sparsity issue. For the purpose of constructing the embedding representation for the segments we decided to employ a matrix factorization approach. A similar technique is also used  in~\cite{deng2016latent,wang2014travel}. Our aim is to discover some latent features that describe
the road segments traffic characteristics.
Each road segment $r_i$ and time window $w$ is  associated with a D-dimensional embedding vector of latent features. Then, the actual travel time report for the segment $r_i$ in the time window $w$ can be approximated by the product of these two latent vectors. In order to satisfy this  condition, the learned embeddings of segments with similar traffic behaviour should be close in the  latent space. 

In order to apply the matrix factorization method we need  to convert the historical reports $R_H$ to a sparse matrix $M \in \mathbb{R}^{N\times W}$, where the $N$ rows  correspond to the  segments $r_i$ of the road network $\mathcal{G}$ and the $W$ columns correspond  to all the time windows of the historical data $w$. Each time window has a size of 30 minutes and each cell corresponds to the average travel time for a segment $r_i$ in this window (i.e. from $10$:$00$ till $10$:$30$), considering all the vehicles that traversed $r_i$. After applying the matrix  factorization the matrix $M$ is decomposed into two matrices $P \in \mathbb{R}^{N\times D}$, $Q \in \mathbb{R}^{W\times D}$ such that $M \approx P \times Q^T = M'$. The rows of $P$ are the D-dimensional embeddings of the road segments and the rows of $Q$ are the D-dimensional embeddings of the time windows.
For estimating the matrices $P$ and $Q$ we minimize the Mean Squared  Error (MSE) between the  original matrix $M$ and the  matrix reconstruction $M'$ using the Stochastic Gradient Descent (SGD). 
The SGD starts with the random matrices $P$ and $Q$ and at each step alters them considering the  direction of the gradient of the objective function. The algorithm terminates when the objective function does not significantly change.




\noindent
{\bf Constructing the Covariance Function:}
\label{sec:covariance_matrix}
Here, we first introduce the properties that characterise the road segments' travel times. Then we describe the complex covariance function that fits our data.
Our aim is to predict the travel time of several queried road segments considering multiple travel time reports which are transmitted by the moving vehicles. Multiple evolving time-series are generated, one for each road segment, and our aim is to make accurate forecasts for their future traffic condition.

To accurately estimate travel time  the following key properties of the time series data should be considered: 
\textit{(i) Periodicity}: the traffic condition of the road segments is periodic in a daily basis, since commuters tend to follow similar trips.
\textit{(ii) Correlation among road segments}:
The information provided by multiple road segments can be  used in order to make predictions  jointly, exploiting the correlations among the road segments and allows us to ameliorate the effects of data sparsity.
\textit{(iii) Short term irregularities}: even if the time series are periodic the traffic condition can be affected by multiple factors (i.e. constructions in the road network, an accident or a social event). This could  generate traffic congestion events that are impossible to detect without monitoring the real-time traffic reports.
\textit{(iv) Noisiness}: the travel time reports are extremely noisy, for instance a driver may be stopped by a traffic light spending 1 minute waiting while another driver may not.

In this work we use Gaussian processes to tackle the travel time estimation problem. We construct appropriately the covariance function providing an excellent fit to the data and characterizing the correlations among the different travel time reports in the process. Here we consider Gaussian processes with a zero mean function. Our goal is to model the travel time ($TT$) of the road segments as a function of the input vector $\textbf{x}\in \mathbb{R}^{D+1}$. $\textbf{x}$ contains the $D$-dimensional embedding representation of the road segment along with the time that the vehicle left the road segment.



\noindent
\begin{minipage}{.48\linewidth}
\begin{equation}
  \bm{x}=
    \begin{bmatrix}
        t,
        \textbf{e}
    \end{bmatrix}^T
\end{equation}
\end{minipage}
\begin{minipage}{.48\linewidth}
\begin{equation}
    \textbf{e}=
    \begin{bmatrix}
        e^1,
        \dots,
        e^D
    \end{bmatrix}
\end{equation}
\end{minipage}

We model the daily variation of the road segments' travel times using a periodic covariance function on the timestamp of measurements $t$, modified  by taking the product with a squared exponential component on \textit{(i)} the timestamp measurement $t$ reducing the impact of older reports and \textit{(ii)} the embeddings in order to reduce the impact of irrelevant road segments.  

\begin{equation}
    k_1(\bm{x},\bm{x}') = \theta_1^2 exp(-\frac{(t-t')^2}{2\theta_2^2} -\frac{(\textbf{e}-\textbf{e}')^T(\textbf{e}-\textbf{e}')}{2\theta_3^2} - \frac{2sin^2(\pi(t-t'))}{\theta_4^2})
\end{equation}


The next term of the covariance function, models the medium term irregularities and the correlations among similar road segments. This term uses a rational quadratic component on the timestamp $t$ and a squared exponential component on the road segments' embeddings $\textbf{e}$. Using this term the travel time prediction of a road segment is affected by the recent reports of road segments with similar traffic behaviour (close in the embedding space), treating the data sparsity problem.

\begin{equation}
    k_2(\bm{x},\bm{x}') = \theta_5^2(1+\frac{(t-t')^2}{2\theta_6 \theta_7})^{-\theta_6} exp(-\frac{(\textbf{e}-\textbf{e}')^T(\textbf{e}-\textbf{e}')}{2\theta_8^2})
\end{equation}

Finally, a noise model is introduced considering the timestamp and the embeddings of the datapoints.

\begin{equation}
    k_3(\bm{x},\bm{x}') = \theta_9^2 exp(-\frac{(e-e')^T(e-e')}{2\theta_{10}^2} - \frac{(t-t')^2}{2\theta_{11}^2})
\end{equation}



The final covariance function is the sum of the previously described covariance functions, $k(\bm{x},\bm{x}') = k_1(\bm{x},\bm{x}') + k_2(\bm{x},\bm{x}') + k_3(\bm{x},\bm{x}')$ , with hyperparameters 
$\bm{\theta}=[\theta_1\dots\theta_{11}]$.
We empirically initialize these hyperparameters
based on our prior beliefs about the data. During the learning procedure
$\bm{\theta}$ is
automatically adjusted in order to appropriately fit the training data. 

\noindent
{\bf Hybrid Travel Time Estimation (HTTE):}
We now present in detail our travel time estimation algorithm for the received query paths. Our algorithm, presented in Algorithm~\ref{alg:htte}, consists of an offline and a real-time stage. The offline stage is responsible to initialize the required variables and is executed only once. The real-time stage of the algorithm receives in a streaming manner query paths and a departure time for each path  and estimates instantly the corresponding travel time considering the already received travel time reports $R_H$.

\noindent\textbf{Offline stage:} Here we perform a set of tasks that initialize our system. Firstly, we compute the average travel time for each road segment for different times of the day, referred as $TT_{avg}$. 
These average travel times will provide us later an approximate estimation of the departure time for each road segment of the given query path. In order to compute $TT_{avg}$ the day is partitioned in time windows of 30 minutes.
Additionally, we compute for each road segment the average travel time and the standard deviation of travel time, referred as $segsStats$. The $segsStats$ variable does not consider the time of the day. These statistics will be used later in order to standardize the travel time reports of the various road segments. Then our algorithm computes an embedding representation $E$ for each road segment as it was described in Section~\ref{sec:embeddings}.
In the next step of the algorithm, a Gaussian process model, with the covariance function described in Section~\ref{sec:covariance_matrix} and zero-mean is trained and its hyperparmeters $\bm{\theta}$ are learned.

Having all the travel time reports that have been received until now  and the road segments statistics and embeddings, our next step is to initialize the Gaussian process models, defined as $gpModels$. In order to avoid generating a large Gaussian process model that would consider the travel time reports for all the city and the whole day we perform spatial and temporal partitioning of $R_H$. This results in multiple $gpModels$ and expedits the estimation of travel times queries.
Each model is affiliated with a particular spatial area and time of the day. When a $gpModel$ is generated, a covariance matrix $K$ is constructed using the covariance function of Section~\ref{sec:covariance_matrix}, and the hyperparameters $\bm{\theta}$. The covariance matrix, for each $gpModel$, describes the correlation between the different travel time reports that have been received till now for a particular time window and area. Since the road segments have different lengths and their travel times deviate significantly, we decided to standardize the travel time reports for each  road segment using the corresponding statistics $segsStats$. Thus the targets $\bm{y}$ for each $gpModel$ are the standardized, with the statistics, travel times and not the raw travel times.

\begin{figure}
\centering
\includegraphics[width=0.48\textwidth]{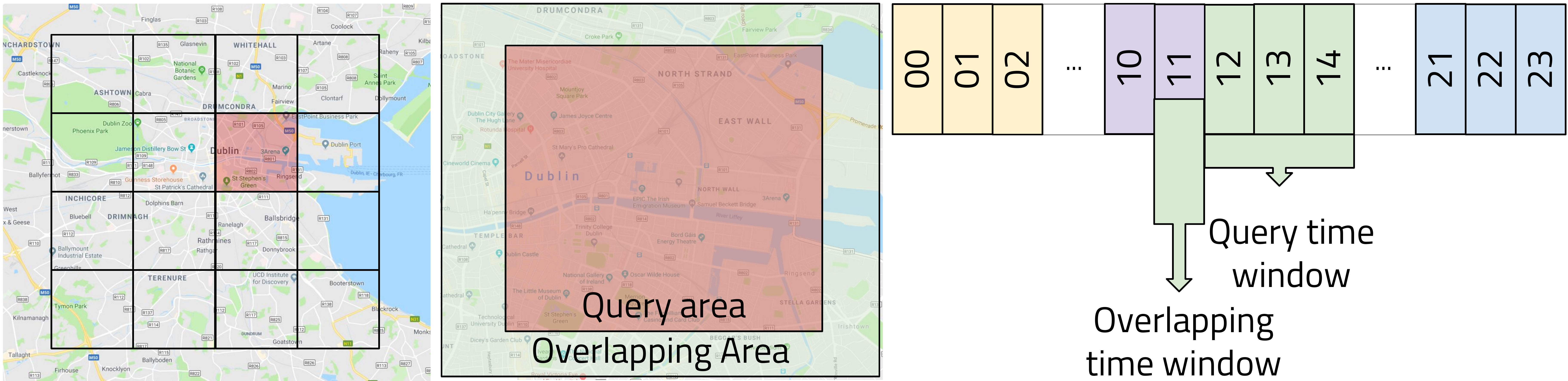}
\caption{Spatial and temporal partitioning.}
\label{fig:spatial-temporal-partitioning}
\end{figure}

The travel time reports are partitioned into different $gpModels$ considering the location of the road segments and their timestamp (Figure~\ref{fig:spatial-temporal-partitioning}). More specifically the whole city is decomposed in smaller areas applying a grid of uniformly sized cells. In order to feed the model we allow spatial overlaps with neighboring spatial grid cells in order to improve the accuracy for the road segments that are near the edges of each cell.
Also, each day is partitioned into smaller time windows. Here we allow temporal overlaps with the previous and the next time windows, providing traffic information at the beginning of each query time window.
The travel time reports that belong in overlapping areas and time windows are inserted into multiple $gpModels$.
Finally, each $gpModel$ contains the recent and historical reports of each query and overlapping time window and area that is associated with.

\noindent\textbf{Real-time stage:} Our system receives query paths in real-time from the $QueriesStream$ and performs instantly the travel time estimation for each given query. 
Initially our algorithm updates the $gpModels$ adding the newly received travel time reports. The update of the $gpModels$ is performed in predetermined periods and not every time a travel time report or a query is received. Such frequent updates would be time consuming. In order to do this we identify the current time window of the day $w$, considering the current time. If the window $w$ has changed from the window  of the previous query $prevW$ then the $gpModels$ are updated.
Each $gpModel$ is updated extending its covariance matrix $K$ with the travel time reports that have been received from the previous update of the $gpModel$ for the investigated spatial area and time window.



The next step of the algorithm is to decompose the given query path in a set of individual road segments ($SubQueries$) and estimate their travel times querying the corresponding $gpModels$. In order to query the $gpModels$ it is required to estimate an approximate departure time $t_i$ for each road segment $r_{q_i}$ of the given query path.
In order to approximate the departure times we begin with the first road segment $r_{q_1}$ of the query path setting as departure time $t_1$ for this road segment the trip's departure time $t_{dep,q}$. Then in order to estimate the departure time for the next road segment $r_{q_i}$ we add to the previous road segment  departure time $t_{i-1}$ the average travel time $TT_{avg}$ of the previous road segment $r_{q_{i-1}}$, as it was computed in the offline stage of the algorithm. This procedure iterates till the last road segment of the query path.
Having an approximate estimation for the departure time for each road segment will allow to perform batch queries to the affected $gpModels$ of the query path, speeding up the execution time of the queries.


Finally, individual queries for the road segments' travel times are posed to the appropriate $gpModel$, considering the spatial and temporal partitioning. The $gpModels$ return standardized travel times, thus the $segsStats$ are required in order to get the actual travel time estimations. The total travel time of the query path $\reallywidehat{TT}_{q}$ is updated considering the estimates of the $gpModels$ for the individual road segments. Finally, $\reallywidehat{TT}_{q}$ is the estimated travel time for the query $q$.



{\SetAlgoNoLine
    \begin{algorithm}[t]
     
     \KwData{$R_H,QueriesStream=[q_1, \dots,q_\infty]$}
     \KwResult{$\reallywidehat{TT}_1,\dots, \reallywidehat{TT}_\infty)$}
     \textbf{Offline Stage}\;
     $TT_{avg} \leftarrow computeAvgTravelTime(R_H)$\;
     $segsStats \leftarrow computeRoadSegmentsStats(R_H)$\;
     $\bm{E} \leftarrow computeEmbeddings(R_H,segsStats)$\;
     
     $\bm{\theta}\leftarrow computeHyperparametersGP(R_H, \bm{E}, segsStats)$\;
     $gpModels \leftarrow initializeMultipleGPs(R_H, \bm{E}, segsStats, \bm{\theta})$\;
     $prevW \leftarrow None$\;
     \textbf{Online Stage}\;
     \ForEach{$q =<P_q,t_{dep,q}>$ \textbf{in} $  QueriesStream$}{
     $\reallywidehat{TT}_{q} \leftarrow 0 $\;
     $w \leftarrow getTimeWindow()$\;
     \If{$w \neq prevW$}{
        $gpModels.update(R_H,segsStats,\bm{\theta})$\;
     }
     $prevW \leftarrow w$\;
     $SubQueries \leftarrow\; decompose\_path(P_q,t_{dep,q}, TT_{avg})$\;
     \ForEach{$<r_i,t_i> \in SubQueries$}{
        $gpModel  \leftarrow findGP(r_i,t_i)$\;
        $\reallywidehat{TT}_{q} \leftarrow \reallywidehat{TT}_{q} + gpModel.query(r_i, t_i, segsStats)$
     }
     }
 \caption{Travel Time Estimation Algorithm}
 \label{alg:htte}
\end{algorithm}
}

\section{Conclusion}
\label{sec:conclusion}
We develop a novel hybrid technique for travel time estimation, that considers recent and historical traffic reports.
An embedding representation for each road segment is learned based on its traffic behaviour. 
This representation is incorporated by a regression technique, handling the data sparsity problem.
This allows our technique to make accurate estimations even if there are no recent traffic reports available for a segment.
Finally, our technique adapts different levels and types of abstraction that allow the real-time travel time estimation.





\bibliographystyle{apa}
\bibliography{references}

\end{document}